\def\year{2021}\relax
\documentclass[letterpaper]{article} 
\usepackage{aaai21}  
\usepackage{times}  
\usepackage{helvet} 
\usepackage{courier}  
\usepackage[hyphens]{url}  
\usepackage{graphicx} 
\usepackage{natbib}  
\usepackage{caption} 
\usepackage{times}
\usepackage{epsfig}
\usepackage{graphicx}
\usepackage{amsmath}
\usepackage{amssymb}
\usepackage{bbm}
\usepackage{bm}
\usepackage{color}
\usepackage{multirow}
\usepackage{array}
\usepackage{booktabs}
\usepackage[ruled]{algorithm2e}
\usepackage{amsmath}
\usepackage[switch]{lineno}

\newcommand{\Tref}[1]{Table~\ref{#1}}

\newcommand{\Fref}[1]{Figure~\ref{#1}}

\newcommand{\etal}{\textit{et al.}}

\newcommand*\samethanks[1][\value{footnote}]{\footnotemark[#1]}

\title{Initiative Defense against Facial Manipulation}
\author{ \Large \textbf{
                        Qidong Huang$^\dagger$, 
                        Jie Zhang$^\dagger$, 
                        Wenbo Zhou\thanks{Corresponding authors, $\dagger$ Equal contribution.}, 
                        Weiming Zhang\samethanks, 
                        Nenghai Yu
                }\\ 
}
\affiliations{
            University of Science and Technology in China\\
            \{hqd0037@mail., welbeckz@, zjzac@mail., zhangwm@, ynh@\}ustc.edu.cn\\
}
\begin{document}

\maketitle

\begin{abstract}
Benefiting from the development of generative adversarial networks (GAN), facial manipulation has achieved significant progress in both academia and industry recently. It inspires an increasing number of entertainment applications but also incurs severe threats to individual privacy and even political security meanwhile. To mitigate such risks, many countermeasures have been proposed. However, the great majority methods are designed in a passive manner, which is to detect whether the facial images or videos are tampered after their wide propagation. These detection-based methods have a fatal limitation, that is, they only work for ex-post forensics but can not prevent the engendering of malicious behavior.

To address the limitation, in this paper, we propose a novel framework of initiative defense to degrade the performance of facial manipulation models controlled by malicious users. The basic idea is to actively inject imperceptible venom into target facial data before manipulation. To this end, we first imitate the target manipulation model with a surrogate model, and then devise a poison perturbation generator to obtain the desired venom. An alternating training strategy are further leveraged to train both the surrogate model and the perturbation generator. Two typical facial manipulation tasks: face attribute editing and face reenactment, are considered in our initiative defense framework. Extensive experiments demonstrate the effectiveness and robustness of our framework in different settings. Finally, we hope this work can shed some light on initiative countermeasures against more adversarial scenarios.

\end{abstract}


\section{Introduction}
With the tremendous success of deep generative models (e.g. conditional-GAN), face manipulation has been an emerging topic in very recent years and a variety of methods \cite{thies2016face2face:,korshunova2017fast,nirkin2019fsgan,natsume2018rsgan,wu2018reenactgan} have been proposed. The face manipulation systems can change the target face with a different attribute, such as hairstyle or expression. As the manipulated results become more and more realistic, these techniques can easily be misused for malicious purposes such as violating individual privacy or even misleading political opinions. In details, the malicious users may edit the portrait image of any person without his/her permission. Moreover, the attacker is able to forge the expression (e.g. lip shape) of political figure's speech video, which might seriously mislead the public.

To alleviate the risks brought by malicious usage of face manipulation, various impressive countermeasures \cite{matern2019exploiting,han2017two,afchar2018mesonet,nguyen2019use,rossler2019faceforensics++,li2020face} have been proposed, most of which are ex-post detection-based methods used for forensics. Notwithstanding the considerably high accuracy in distinguishing forged facial images or video from the real ones, these methods are too passive and hard to eliminate the damages caused by malicious face manipulation, because the generation and wide spread of such forgeries have already become a fait accompli. How to proactively prevent such threats before its happening is a significant but seriously under-researched problem. Very recently, Ruiz \etal \cite{ruiz2020disrupting} propose a gradient-based method to attack facial manipulation models, which is regarded as the baseline method of this paper. However, its white-box assumption is not practical. The baseline method needs the inner information about the target models. Besides, it is invalid to some other types of manipulation tasks, like the real-time face reenactment. 


In this work, we propose the concept of initiative defense against face manipulation, which is a totally fresh angle of defense. Specifically, we leverage a poison perturbation generator $\mathbf{PG}$ to generate invisible perturbation, and then superpose it onto the original face data before releasing the data to the public. The forger can only have access to the infected face data, and due to the existence of poison perturbation, the performance of face forgery models will be significantly degraded, whether infected data is used in inference stage or in training process. In other words, the infected face data must satisfy two ultimate objectives: (1) is visually consistent to the clean one and (2) can degrade the performance of the target model greatly.


To achieve these goals, we devise a two-stage training framework of initiative defense, which is generally applicable to different facial manipulation tasks and different adversarial settings. It should be pointed out that the baseline method methoned above can be regarded as a special example of our framework. Here, we take the defense against attribute editing task as an example to elaborate our framework briefly. As shown in \Fref{fig:pipeline_stargan}, in the stage $A$, we train a surrogate model $\mathbf{SM}$ to imitate the target manipulation model's behavior, while we aim to obtain the corresponding perturbation  generator $\mathbf{PG}$ in the stage $B$. In practice, it is hard to train a $\mathbf{PG}$  with the well-trained $\mathbf{SM}$ in the stage $A$. 
The main reason is that, the infected data produced by the initial $\mathbf{PG}$ will easily solve the objective (2) mentioned above, and make the training process trapped in the local optimum because of the non-convex property of DNNs. To this end, we propose an alternating training strategy to train $\mathbf{SM}$ and $\mathbf{PG}$ step by step. The whole training procedure is illustrated in Algorithm 1. Different from the popular adversarial training approach, in our framework, only the update of perturbation generator $\mathbf{PG}$ is influenced by the surrogate model $\mathbf{SM}$ while the $\mathbf{SM}$ is regularly updated in an isolated manner. For different tasks, we further leverage some task-specific strategies to enhance the defense effectiveness.

We conduct experiments on two typical types of face manipulation tasks: face attribute editing and face reenactment. Extensive experiments demonstrate the effectiveness of the proposed framework and the robustness in different adversarial settings, even in a black-box way. Furthermore, some ablation analysis is provided to justify the motivation of our design.

To summarize, the contributions of this paper are threefold as below:

\begin{itemize}
    \item We introduce the concept of initiative defense against face manipulation, which is a new perspective of countermeasures. And we hope it can inspire more great works for this seriously under-researched field.
    \item We propose the two-stage training framework, which is generally applicable to different face manipulation tasks. Furthermore, we leverage an alternating training strategy to achieve the designed objectives, and some task-specific strategies to enhance the defense performance.
    \item Extensive experiments demonstrate the proposed framework can resist the manipulation from the forgers in different settings, including the black-box adversarial scenarios.
    
\end{itemize}

\section{Related work}
\noindent\textbf{Deep Face Forgery.} 
Deep face forgery techniques have achieved tremendous progress in the past few years. According to different forgery goals, existing methods can be roughly divided into two categories: \textit{face swapping} and \textit{face manipulation}. Face swapping is a representative framework in deep face forgery which swaps the face between a source and a target person. In this class of methods \cite{deepfakes,faceswap,bao2018towards,afchar2018mesonet}, auto-encoder and GANs are commonly used as backbones to generate realistic results. All these methods are ``identity-changed'' which focus on exchanging the identity face region and seldom concerns other attributes out of the face region. Face manipulation is a more common type of deep face forgery methods, this type of methods \cite{choi2018stargan,thies2016face2face:,Pumarola_ijcv2019,kim2018deep} includes many cases such as face attribute editing which modifies part of the facial attributes (e.g., hair style), and face reenactment which transfers the facial expression from one person to another. StarGAN \cite{choi2018stargan} achieves facial attribute transfer among multiple domains by using a single generator. It is adequately trained on public face datasets and edit the target attribute in the inference stage. Face2face \cite{thies2016face2face:} is a typical approach of face reenactment. To transfer the facial expression from source to target, it requires a considerable amount of the target face data to train the generative model for each specific person.

Usually, the realism of results generated by face swapping are quite limit especially when source and target attribute are very different. However, face manipulation is easier to achieve much better visual quality because that it does not need to change the original face. Therefore, face manipulation has a wider range of application scenarios than face swapping currently. In this work, we mainly focus on facial manipulation to implement our initiative defense. 
\noindent\textbf{Existing Countermeasures against Facial Manipulation.} As face manipulation causes great threaten to individual privacy even political security, it is of paramount importance to develop countermeasures against it. Previous works \cite{rossler2019faceforensics++,matern2019exploiting,li2020face,qian2020thinking,dang2020detection} are commonly designed in a passive manner. For instance, Rossler \etal \cite{rossler2019faceforensics++} introduce a simple but effective Xception Net as a binary classifier for face manipulation detection. Face X-ray \cite{li2020face} considers the detection from a different view which focuses on the blending artifacts rather than the manipulation artifacts. The blending operator could not be avoided in the process of manipulation, thus face X-ray obtains better results than most works. However, although some detection-based methods achieve considerably high accuracy in ex-post forensics, they still can not eliminate the influences of malicious applications which had been caused in the wide spreading.

Recently, the countermeasures which aim to prevent malicious behavior before the manipulation have been studied. For example, Fawkes \cite{shan2020fawkes} protects faces from being stolen by recognition systems. But it can not prevent private face images from being edited by malicious users. Ruiz \etal \cite{ruiz2020disrupting} propose a gradient-based method to attack facial manipulation systems in white box successfully, which protects faces from being manipulated by some specific models. However, this method is time-consuming and laborious, which requires optimization for each single image in each iterations and performs badly in both gray-box and black-box scenes. To address these limitations, this paper is the first try to establish a complete framework of initiative defense against face manipulation, which aims to be more efficient and robust.

\section{Method}
In this section,  we will elaborate the proposed framework of initiative defense in detail. Before that, the formal problem definition will be introduced firstly.


\noindent\textbf{Problem Definition.}
For facial manipulation tasks, we roughly divided them into two categories: \textit{model-based manipulation} and \textit{data-based manipulation}.  The corresponding example  tasks are facial attribute editing and face reenactment, respectively. Given a face image  or video $x$,  the former type is to modify  $x$ to forged example $y$  with a well-trained manipulation model $\mathcal{M}$, while the latter one is to utilize the given face data $x$ to train a new model $\mathcal{M}_x$, and feed some guidance information $z$ (e.g. landmark) into the model to generate modified face $y$. Therefore, the target of facial manipulation tasks can be formulated by
 \begin{equation} \label{1}
     y=\left\{
\begin{aligned}
  & \mathcal{M}(x), & \text{model-based} ;\\
  & \mathcal{M}_x(z), & \text{data-based} .  
\end{aligned}
\right.
 \end{equation}
 Before the forgers get access to the face data $x$, we can pre-process it to obtain the infected face data $x'$ in an additive way, i.e.,
  \begin{equation}
     x' = x + \epsilon \cdot \delta_x, \qquad \lVert \delta_x \rVert_\infty \leq \tau.
 \end{equation}
 Here, $\delta_x$ denotes the perturbation superposed on $x$. $\epsilon$ represents an adjustable intensity of  $\delta_x$, and  $\tau$ is an  threshold to constrain $\delta_x$.
With the infected data $x'$, the forger will acquire the corresponding infected forged example $y'$.
 One of our objective is to guarantee the visual consistence between clean data $x$ and infected data $x'$.  The other objective is to destroy the generation of $y$, namely that, to maximize the distance $\mathcal{D}$ between the clean forged face $y$ and the infected forged face $y'$, i.e.,
 \begin{equation} \label{3}
     \delta_x = \arg \max_{\delta} \mathcal{D}(y,  y').
 \end{equation}

\begin{algorithm}[h]
    \caption{Two-stage Training Framework}
    \LinesNumbered
    \KwIn{number of iterations $maxiter$, original training dataset $D$.}
    \KwOut{a perturbation generator $PG_{best}$.}
    $//$ Initialization\;
    $SM_0, PG_0 \gets RandomInit()$ \;
    $PG_{bset} \gets PG_0$,  $maxdist \gets -inf$ \;
    \For{$i = 1$ $\mathbf{to}$ $maxiter$}{
        $x \sim SelectMiniBatch(D)$\;
        $//$ \textbf{Stage A}: train $SM$ on clean data $x$ regularly\;
        $SM_i \gets Update(SM_{i-1}, x, \mathcal{L}_{A})$\;
        $//$ Get clean forgery $y$ based on Eq.\eqref{1}\;
        $y \gets GetForgery(x)$\;
        $//$ \textbf{Stage B}: obtain $x'$ using last iteration's $PG$\;
        $x' \gets x + \epsilon \cdot PG_{i-1}(x)$\;
        $//$ Acquire the corresponding $y'$  based on Eq.\eqref{1}\;
        $y' \gets GetForgery(x')$\;
        $//$ Design task-specific influence loss $\mathcal{L}_{A \rightarrow B}$ to \\
        $//$ maximize the distance between $y$ and $y'$\;
        $\mathcal{L}_{A \rightarrow B} \gets \max \mathcal{D} (y, y')$\;
        $//$ Train $PG$ based on $\mathcal{L}_{A \rightarrow B}$ and  $\mathcal{L}_{B}$\;
        $PG_i \gets Update(PG_{i-1}, x, \mathcal{L}_{B}, \mathcal{L}_{A \rightarrow B})$\;
        $//$ Compare distances and select the best one\;
        \If{$maxdist < \mathcal{D} (y, y')$}{
            $maxdist \gets \mathcal{D} (y, y')$ \;
            $PG_{best} \gets PG_{i}$ \;
        }
    }
    \Return $PG_{best}$
    \label{alg:Algo}
\end{algorithm}
\noindent\textbf{Two-stage Training Framework of Initiative Defense.}
For the baseline method \cite{ruiz2020disrupting}, its assumption that having the full access to the target manipulation model is not quite practical.  To this end, we substitute  a surrogate model $\mathbf{SM}$  for the target manipulation model $\mathbf{M}$, which is feasible because of the accessibility  to the model type and training procedure of a specific manipulation task. Besides, we devise a poison perturbation generator $\mathbf{PG}$ to generate the $\delta_x$ =  $\mathbf{PG}(x)$, which will incur minor computational cost compared to the  gradient-based method adopted by Ruiz \etal. We propose the two-stage training framework  to train $\mathbf{SM}$ and $\mathbf{PG}$. Intuitively, we attempt to train the $\mathbf{PG}$ with a well-trained $\mathbf{SM}$, but it does not work because the objective function described in Eq.\eqref{3} will be fulfilled  at the very beginning of the $\mathbf{PG}$'s training, which makes it trapped into the local optimum. Therefore, we propose the  alternating training strategy to train both  the $\mathbf{SM}$ and $\mathbf{PG}$ one by one from scratch. Concretely, in every training iteration, we first update the surrogate model $\mathbf{SM}$ regularly and get the corresponding infected forgery $y$ in the stage A. And in the stage B, with the infected data $x'$ calculated by the  trained $\mathbf{PG}$ in the last iteration, we obtain the infected forgery $y'$ based on Eq. \eqref{1}. And then, a task-specific influence loss function $\mathcal{L}_{A \rightarrow B}$ will be designed to feedback the training influence from the stage $A$ to the stage $B$. Both $\mathcal{L}_{A \rightarrow B}$ and the conventional loss objective $\mathcal{L}_{B}$ are leveraged to update the $\mathbf{PG}$ in the stage $B$. The pseudo code of two-stage training framework is illustrated in Algorithm 1.



%

\begin{figure*}[!h]
\centering
\includegraphics[width=1.0\linewidth]{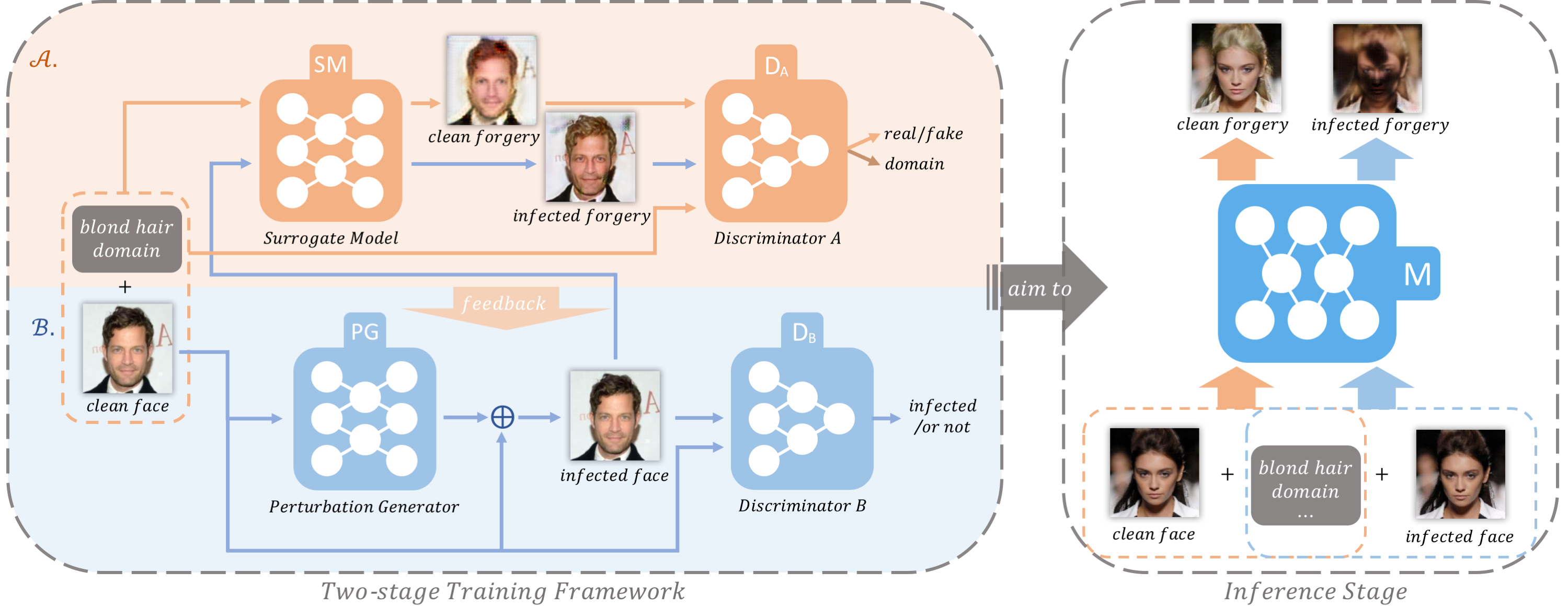}
\caption{The overall pipeline to defend facial attribute editing. We set the blond hair as the target domain. The left part is two-stage training framework, and in the stage A, we train a surrogate model $\mathbf{SM}$ regularly on clean faces and domain labels to imitate the behavior of the target model $\mathbf{M}$. In the stage B,  a perturbation generator $\mathbf{PG}$ is trained on clean faces and constrained with the extra influence loss $\mathcal{L}_{A \rightarrow B}$ feedbacked from the stage A.  During the inference stage, the infected faces generated by the well-trained $\mathbf{PG}$ will disrupt the target model $\mathbf{M}$ severely.}
\label{fig:pipeline_stargan}
\end{figure*}

\noindent\textbf{Defending Against Model-based Manipulation.}
As shown in \Fref{fig:pipeline_stargan}, we take the facial attribute editing task as example to elaborate the network architectures and loss functions adopted in our two-stage training framework in detail.
We adopt an auto-encoder like networks with 6 residual blocks (`` Res6'')  and UNet-128 \cite{ronneberger2015u-net:} as the default architecture of surrogate model $\mathbf{SM}$ and  perturbation generator $\mathbf{PG}$, respectively. Two discriminators $\mathbf{D_A}$ and $\mathbf{D_B}$  are introduced in both stages, which are common convolutional networks (7 convolutional layers appended with one dense layer). For $\mathbf{SM}$ in the stage $A$, we follow the settings in \cite{choi2018stargan} to train it step by step.  The objective loss function to train $\mathbf{PG}$ consists of two parts: the influence loss $\mathcal{L}_{A \rightarrow B}$ and the conventional training  loss  $\mathcal{L}_{B}$, i.e.,
\begin{equation} \label{4}
    \mathcal{L}_{PG} =\mathcal{L}_{A \rightarrow B}  + \lambda \cdot  \mathcal{L}_B ,
\end{equation}
where $\lambda$ is the hyper parameter to balance these two loss terms. Below we will introduce the detailed formulation of  $\mathcal{L}_{A \rightarrow B}$ and $\mathcal{L}_{B}$, respectively. Here, only the adversarial loss  $\mathcal{L}_{adv}$ is considered in $\mathcal{L}_B$, which is enough to guarantee the visual quality between $x$ and $x'$, i.e.,  
\begin{equation}
    \begin{split}
        \mathcal{L}_{adv} = \mathbb{E}_x [D_B(x)] - \mathbb{E}_{x'} [D_B(x')] \\
        - \lambda_{1} \cdot \mathbb{E}_{\hat{x}} \Big[ ( \lVert \nabla_{\hat{x}} D_B(\hat{x}) \rVert_2 - 1)^2 \Big],
    \end{split}
\end{equation}
where the last term leverages Wasserstein distance to penalize the gradient for training stability \cite{arjovsky2017wasserstein,gulrajani2017improved} and $\lambda_{1}$ is the hyperparameters indicating the weights to the corresponding term. The responsibility of $\mathcal{L}_{A \rightarrow B}$ is to maximize the distance between the clean forgery $y$ and infected forgery $y'$. In this task,  $\mathcal{L}_{A \rightarrow B}$ is defined as the weighted sum of three terms, i.e.,
 \begin{equation}
    \mathcal{L}_{A \rightarrow B} =  \lambda_{2} \cdot \mathcal{L}_{bs} + \lambda_{3} \cdot \mathcal{L}_{cyc} + \lambda_{4} \cdot \mathcal{L}_{dom} .
\end{equation}
where the first two terms are to degrade the visual quality of $y'$ compared with $y$ in diverse domains and the last term is to confuse the discriminator $\mathbf{D_A}$ to classify $y'$ as fake outputs and the farthest domain from the one of $y$. In detail, the basic loss $\mathcal{L}_{bs}$ is used to maximize the pixel-level difference between $y'$ and $y$, i.e.,
\begin{equation}
    \mathcal{L}_{bs} = \mathbb{E}_{x, x', c} \Big[ -\sum_j \mu_j \cdot \lVert SM(x, c_j) - SM(x', c_j)\rVert_1 \Big],
\end{equation}
where $c_j$ denotes to a series of target domains derived from the original domain $c$ of $x$, and $ \lVert \cdot  \rVert_1$ is $L_1$ norm distance.
Here, $\mu_j$ is to balance the influence degree among different domains due to the  discrepant area ratio in pixel-level, and we achieve it in an indirect way, i.e.,
\begin{equation}
    \mu_j = \frac{\lVert x - SM(x, c_j) \rVert_1}{\sum_j \lVert x - SM(x, c_j) \rVert_1}.
\end{equation}
Cycle-consistency loss is popular for many un-supervised tasks and we devise $\mathcal{L}_{cyc}$ to disrupt the consistency, which is described as  
\begin{equation}
    \mathcal{L}_{cyc} = \mathbb{E}_{x', c} \Big[ -\sum_j \mu_j \cdot \lVert x' - SM(SM(x', c_j), c)\rVert_1 \Big].
\end{equation}
To enlarge the difference of $y$ and $y'$ in high-dimensional level, we calculate the inverse domain $c_{rj}$ for each $c_j$ as the farthest domain, and minimum the confidence probability of $y'$ from the perspective of $\mathbf{D_A}$ meanwhile. Therefore, the domain loss $\mathcal{L}_{dom}$ is defined as
\begin{equation}
    \begin{split}
        \mathcal{L}_{dom} = \mathbb{E}_{x', c} \Big[ \mathbb{E}_{c_j}[- \log D_A(c_{rj}\vert SM(x', c_j))] \Big] \\
        + \mathbb{E}_{x', c} \Big[ \mathbb{E}_{c_j}[D_A(SM(x', c_j))] \Big].
    \end{split}
\end{equation}

\begin{figure}[!h]
\centering
\includegraphics[width=1.0\linewidth]{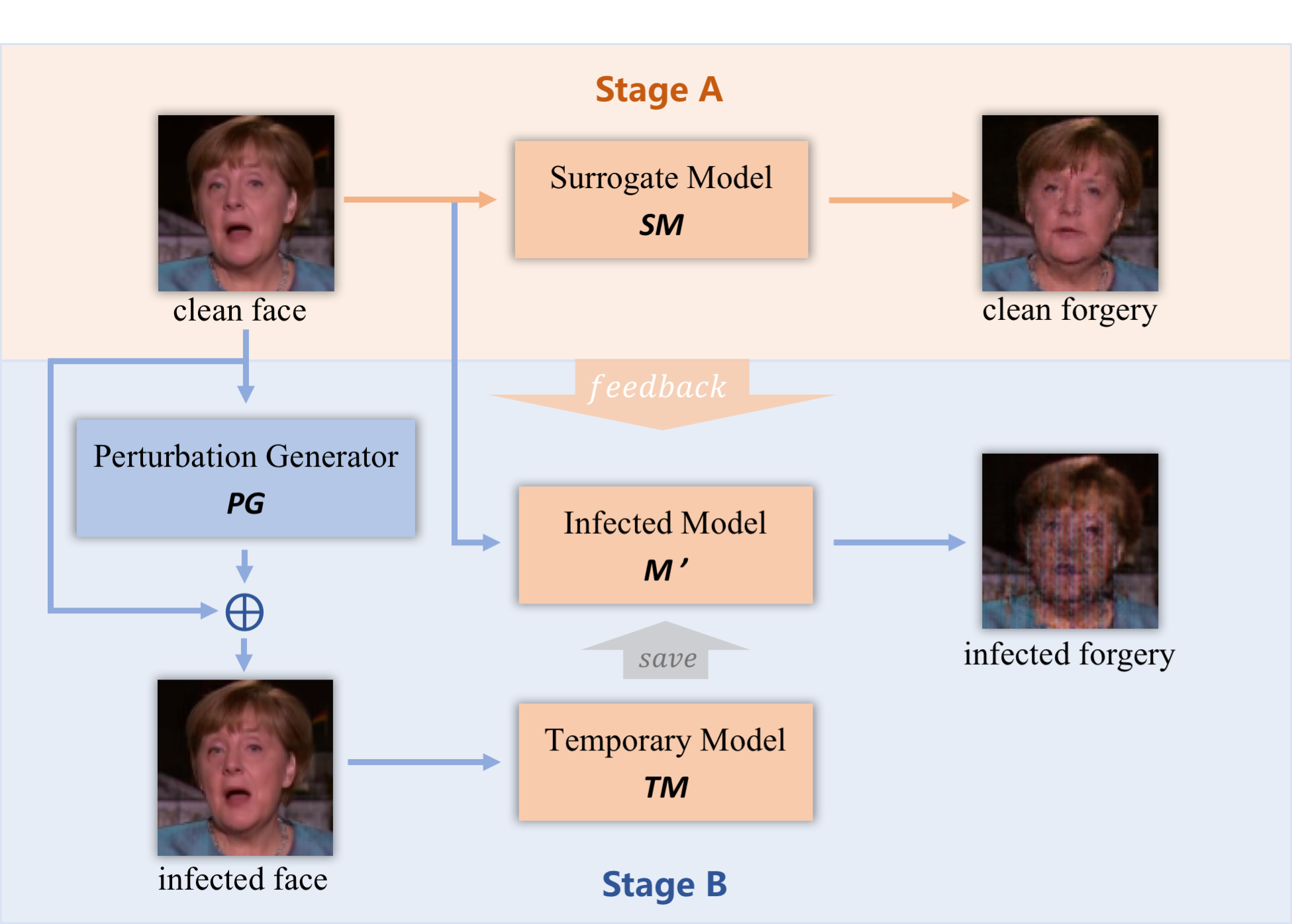}
\caption{The  brief pipeline for face reenactment task. We shall point out that we utilize a temporary model $\mathbf{TM}$ to maintain the gradients of $\mathbf{PG}$, while $\mathbf{TM}$'s parameters is copied into $\mathbf{M}$ to calculate the  influence loss $\mathcal{L}_{A \rightarrow B}$.}
\label{fig:pipeline_face2face}
\end{figure}

\vspace{0.5em}
\noindent\textbf{Defending against  Data-based Manipulation.}
Face reenactment is the classic type task of data-based manipulation and the defending against Face2Face  is illustrated in \Fref{fig:pipeline_face2face} for instance.  Similarly, a surrogate model $\mathbf{SM}$ is learned conventionally according to the guidance by \cite{thies2016face2face:} in the stage A, which can be regarded as an image-to-image translation model. In the stage B, $\mathcal{L}_{adv}$ mentioned above is also applied in this task as $\mathcal{L}_{B}$ for $\mathbf{PG}$'s training. 
The difference is the influence loss $\mathcal{L}_{A \rightarrow B}$ (in Eq.\eqref{4})  here designed to greatly weaken the ability of infected model $\mathbf{M'}$ compared with surrogate model $\mathbf{SM}$, i.e.,
\begin{equation} \label{11}
    \mathcal{L}_{A \rightarrow B} =  \mathbb{E}_{x, z} \Big[ \lVert (SM_x(z) - x) \rVert_1 - \lVert (M_{x'}'(z) - x) \rVert_1 \Big],
\end{equation}
in which we simply use the  $\lVert (SM_x(z) - x) \rVert_1$  to represent the reconstruction ability of the image translation model $\mathbf{SM_x}$. Intuitively, it is equivalent to a distance penalty to further corrupt the performance of the infected model $\mathbf{M_{x'}'}$. Besides, as shown in \Fref{fig:pipeline_face2face}, a temporary model $\mathbf{TM}$ is introduced to maintain the gradient information to $\mathbf{PG}$, and we copy its parameters into an infected model $\mathbf{M'}$ to compute the influence loss $\mathcal{L}_{A \rightarrow B}$. Similar training mechanism is also leveraged in \cite{feng2019learning}. To simplify the framework, we assume that the perturbation will not influence the landmark of face data, which is perhaps a more rigorous assumption for this problem.

\noindent\textbf{Task-specific Defense Enhancement.}
To further boost the defense ability, we also leverage some training skills to enhance the defense ability for specific task.
For facial attribute editing task, we integrate the extra influence loss $\mathcal{L'}_{A \rightarrow B}$, calculated between the current $\mathbf{PG}$ and the $\mathbf{SM}$ in last iteration, into the original objective loss $\mathcal{L}_{PG}$ illustrated in Eq.\eqref{4}, i.e.,
\begin{equation}
    \mathcal{L}_{PG} = \mathcal{L}_{A \rightarrow B} + \mathcal{L'}_{A \rightarrow B} + \lambda \cdot \mathcal{L}_B,
\end{equation}
which will absorb more influence information from the stage A  and stabilize the optimization.

As for Face2Face task, we introduce the attention-based facial mask $m(x)$ of clean data $x$ into Eq.\eqref{11}, namely that, $\lVert (SM_x(z) - x) \cdot m(x) \rVert_1 - \lVert (M_{x'}'(z) - x) \cdot m(x) \rVert_1$, which assigns the facial area as 1.0 and the rest as 0.01. Guided by BiSeNet-based face matting method \cite{yu2018bisenet:}, $m(x)$ can enable the optimization to pay more attention to corrupted facial area.


\section{Experiments}
In this paper, we conduct experiments on two classical manipulation tasks: facial attribute editing and face reenactment. To demonstrate the effectiveness and robustness of our method, we first show the newly introduced initiative defense framework can greatly disrupt the malicious manipulation models while guaranteeing the visual quality of pre-processed face data. Then, we verify the proposed method is robust in different adversarial settings. Finally, some ablation studies are provided to justify the motivation of the training strategies we leverage and prove the feasibility of the extensions to the  combined manipulation scenario.  

\vspace{0.5em}
\noindent\textbf{Implementation details.}  For facial attribute editing, we use face images from the CelebA dataset \cite{liu2015faceattributes}, which is further split into 100000 images for two-stage training, 100000 for training the target model and 2600 for inference stage. As for face reenactment, Face2Face is investigated as the representative forgery model, with the famous Ms Merkel's speech video as a face data example. We samples 1500 frames in this video and shuffle them randomly, which are utilized in the training phase as shown in \Fref{fig:pipeline_face2face}. Then we obtain the final infected model $\mathbf{M'}$ trained on the whole infected video (15000 frames), which is poisoned by the well-trained  $\mathbf{PG}$. And another 100 clean faces with expression landmark information are prepared for testing.

In the first scenario, the learning rate of both $\mathbf{PG}$ and the surrogate model are assigned as 0.0001 by default, and five randomly selected attribute domains are involved in the two-stage training framework. Empirically, we adopt $\lambda = 0.01$, $\lambda_{1}, \lambda_{2} = 10$, $\lambda_{3} = 2.5$, and  $\lambda_{4} = 1$ in the related loss function. 
For the second scenario, the initial learning rates of all models are equal to 0.0002 and $\lambda = 0.01$ by default.


\vspace{0.5em}
\noindent\textbf{Evaluation Metrics.} 
To evaluate the visual quality, PSNR is used by default.
To investigate the defending effect under each condition, we compute the average $L_1$ and $L_2$ norm distance, LPIPS \cite{zhang2018perceptual} and perceptual loss \cite{Johnson2016Perceptual} between  forgeries and original faces to measure the degradation degree of forgery generation. Specifically, in the first scenario (attribute editing), we regard the defense as success if the $L_2$ distance is bigger than 0.05. Based on it, the defense success rate (DSR) is further defined as the ratio of poisoned face images whose corresponding forgery is successfully corrupted. 
While in the second scenario (Face2Face), the corrupted area is more sparse than that in attribute editing task, and therefore, we replace the $L_2$ norm  in the calculation of DSR by the sparse metric $L_1$ norm. 
In order to show the effectiveness of the proposed defense more intuitively, we adopt Local Binary Pattern (LBP)  to depict the texture characteristics and identity information  of the corresponding faces, which is  often leveraged in some traditional face recognition algorithms \cite{guo2010a,zhang2010local}.

\begin{figure}
    \centering
    \includegraphics[width=1.0\linewidth]{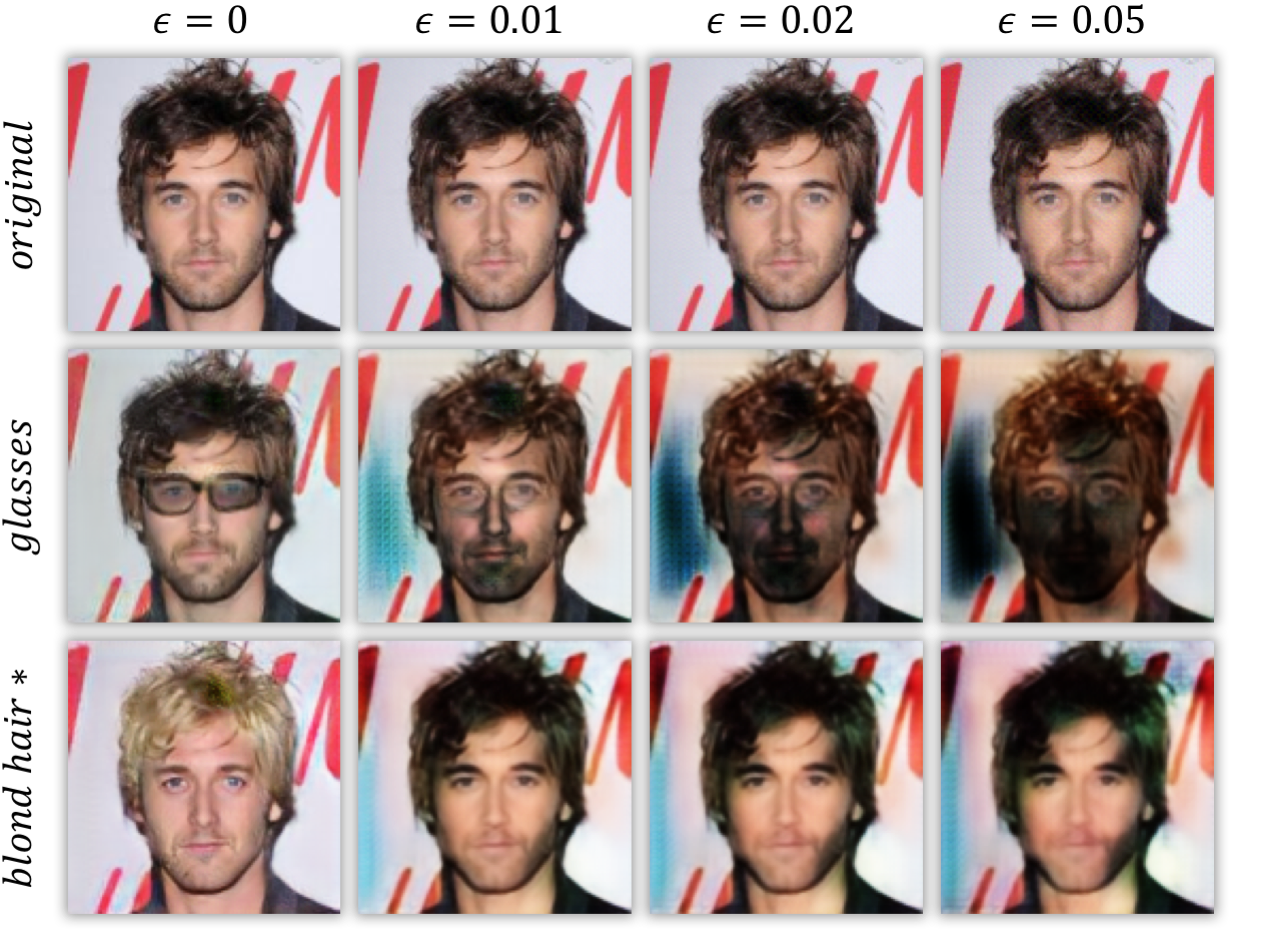}
    \caption{Some visual examples of defending against attribute editing under different perturbation intensities $\epsilon$. The first row are the infected data and the last two rows are the infected forgery with different domains. $\ast$ denotes the wild domain not involved in the training phase of  $\mathbf{PG}$.}
    \label{fig:stargan_show_1}
\end{figure}

\begin{figure*}[h]
\begin{minipage}{0.32\linewidth}
    \centering
    \includegraphics[width=1\linewidth]{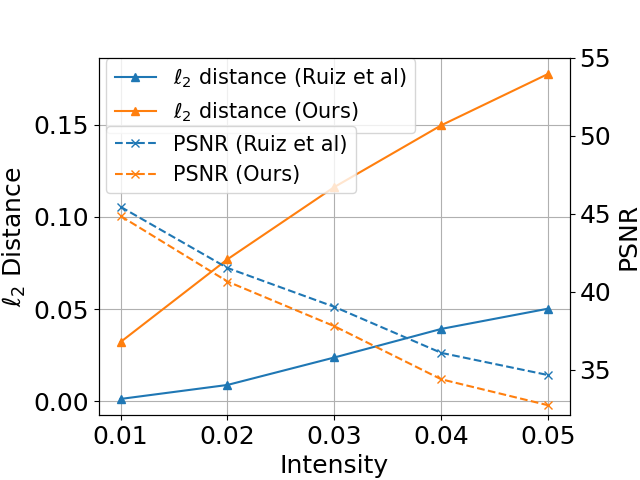}
\end{minipage}
\hfill
\begin{minipage}{0.32\linewidth}
    \centering
    \includegraphics[width=1\linewidth]{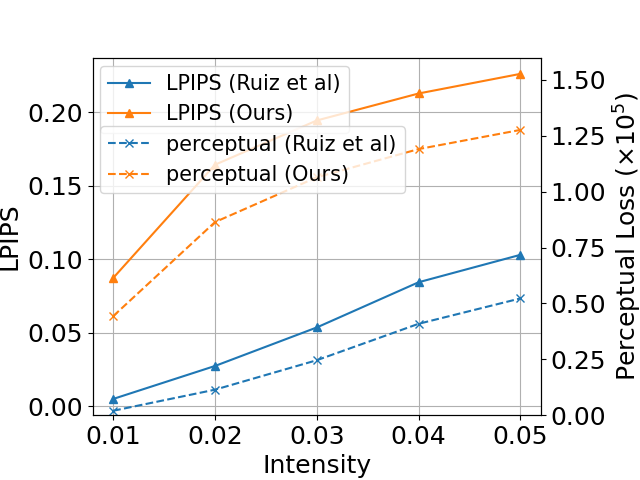}
\end{minipage}
\hfill
\begin{minipage}{0.32\linewidth}
    \centering
    \includegraphics[width=1\linewidth]{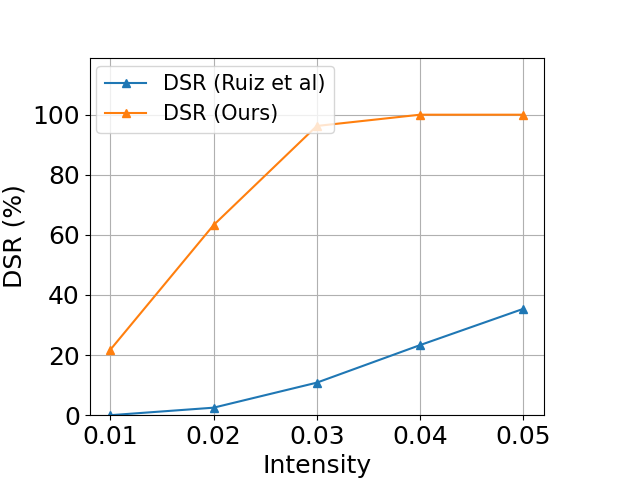}
\end{minipage}
\caption{Quantitative comparison of defending against attribute editing between the baseline method \cite{ruiz2020disrupting} and our approach under different perturbation intensities $\epsilon$. \textbf{Left}: PSNR for infected data and average $L_2$ for infected forgery; \textbf{Middle}: LPIPS and perceptual loss for infected forgery; \textbf{Right}: Defendse success rate (DSR).}
\label{fig:graph1}
\vspace{-0.5em}
\end{figure*}

\begin{figure}[!h]
    \centering
    \includegraphics[width=1.0\linewidth]{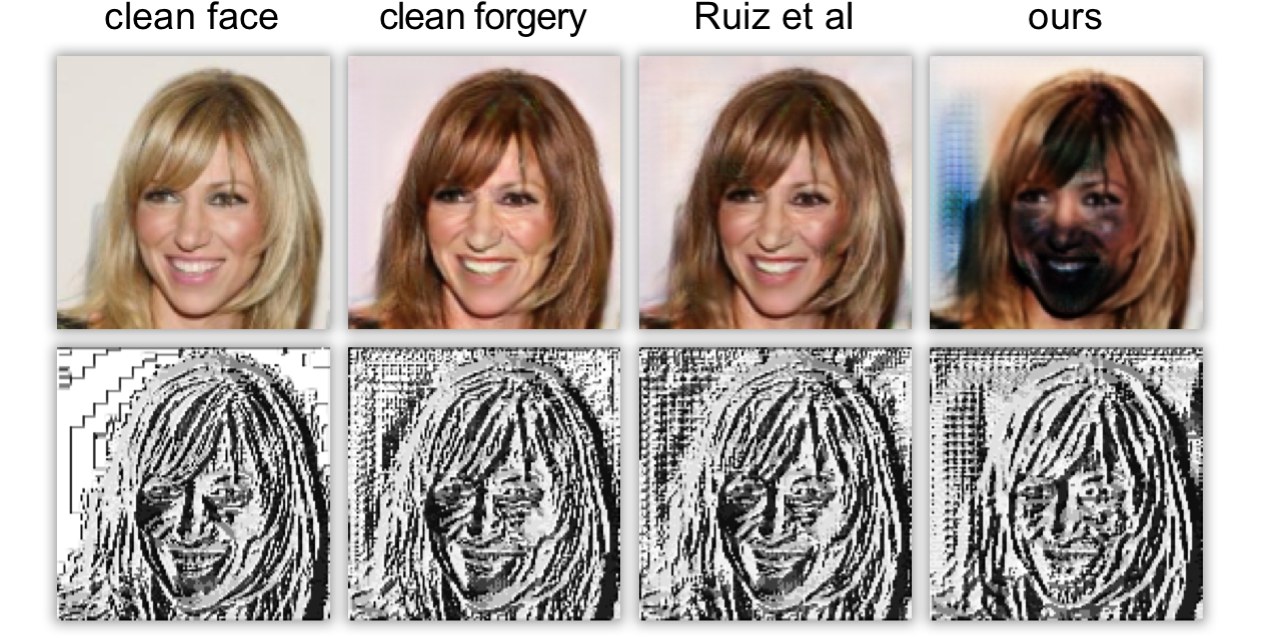}
    \caption{Visual comparison of defending against attribute editing between the baseline method \cite{ruiz2020disrupting} and our method in pixel-level (top) and LBP-level (bottom). ($\epsilon = 0.03$, in gray-box way.)}
    \label{fig:stargan_show_2}
    \vspace{-0.5em}
\end{figure}

\vspace{0.5em}
\noindent\textbf{Effectiveness of The Proposed Initiative Defense.} To demonstrate the effectiveness of the  proposed initiative defense, we conduct the basic experiments in gray-box way, that is,  we are  clear about the network architecture and  domain information (especially for facial attribute editing task) of the attacker's manipulation model $\mathbf{M}$, but its inner information such as  parameters and gradients is still unknown for us. 
	Then, we adopt the same network structure and attribute domain to train the surrogate model $\mathbf{SM}$.

For face attribute editing task, our method can greatly disrupt the malicious manipulation model while guaranteeing the visual quality of infected data. In addition, a control experiment on  different perturbation intensities $\epsilon$ are considered. Some visual examples are shown in \Fref{fig:stargan_show_1}, and we can observe the damaged scale on face forgery becomes larger as the increase of $\epsilon$, which means the defending effect gets better. In other words, the forger can not manipulate our infected face to his/her desired domain, even if the threshold $\epsilon$ is set as 0.01, which represents very slight perturbation. 
Furthermore, we compare the effectiveness of our approach with a gradient-based method \cite{ruiz2020disrupting} proposed recently, which optimizes every single face image separately in white-box way. As shown in \Fref{fig:graph1} and \Fref{fig:stargan_show_2}, we can achieve great incremental defense effectiveness while keeping a very comparable visual quality.  

\begin{figure}
    \centering
    \includegraphics[width=1.0\linewidth]{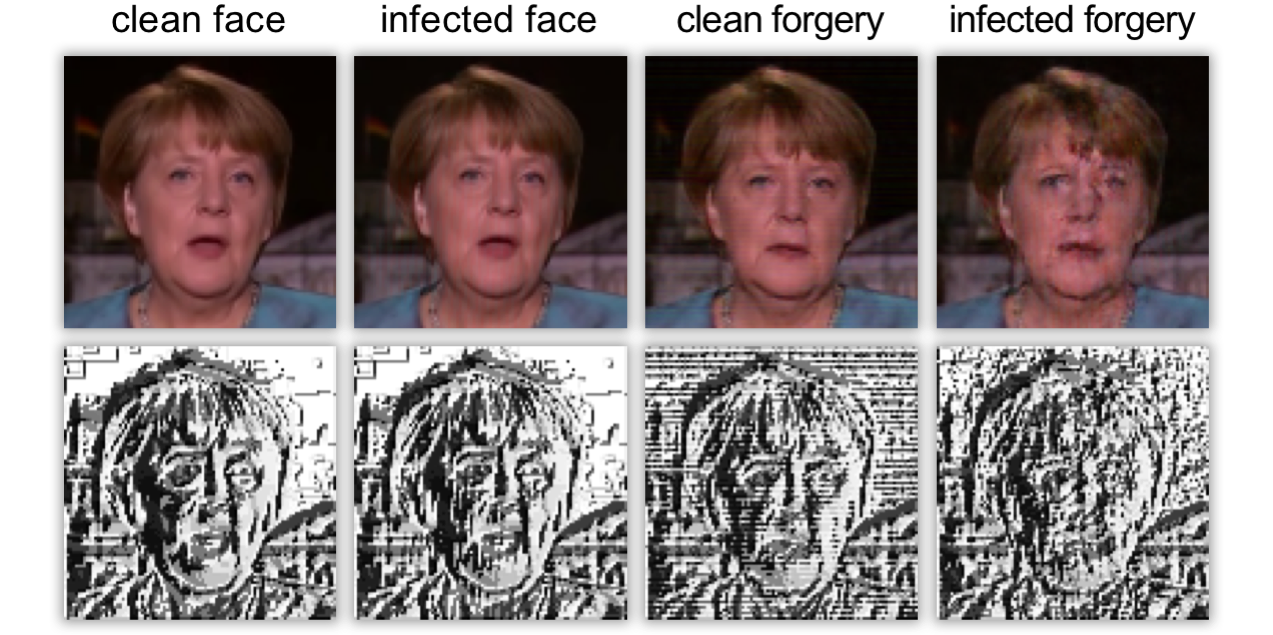}
    \caption{Some visual examples of defending against face reenactment ($\epsilon=0.02$). The bottom row is the  pertinent LBP features of the top row, and the forgeries are generated from the same  expression landmark.}
    \label{fig:face2face_show_1}
    \vspace{-0.5em}
\end{figure}

As for the experiments of defending against Face2Face,  we obtain target manipulation  model $\mathbf{M}$ and infected model $\mathbf{M'}$  by training  with clean video and poisoned video data, and then evaluate them on the same face landmarks. In \Fref{fig:face2face_show_1}, the output of the both visual quality and texture characteristic are corrupted significantly after data poisoning, thus the forger is unable to successfully manipulate the infected video protected by particular $\mathbf{PG}$. We shall point out that the baseline method is not applicable to  data-based manipulation, and to the best of our knowledge, the proposed method is the first attempt to such adversarial scenario. 

\begin{table}[t]
    \scriptsize
    \centering
    \setlength{\tabcolsep}{1mm}{
    \begin{tabular}{ccp{8.5mm}<{\centering}p{8.5mm}<{\centering}p{8.5mm}<{\centering}p{8.5mm}<{\centering}p{8.5mm}<{\centering}p{8.5mm}<{\centering}}
        \toprule
        \multirow{2}{*}{Domain} & \multirow{2}{*}{Setting} & \multicolumn{2}{p{17mm}<{\centering}}{Ruiz $\etal$}
        & \multicolumn{2}{p{17mm}<{\centering}}{Ours} & \multicolumn{2}{p{17mm}<{\centering}}{Ours$\dag$}\\
        \cmidrule(lr){3-4}\cmidrule(lr){5-6}\cmidrule(lr){7-8}
        & & DSR & $L_2$ & DSR & $L_2$ & DSR & $L_2$\\
        \midrule
        \multirow{4}{*}{SD} & Res6* & 35\% & 0.050 & \textbf{100\%} & \textbf{0.177} & 95\% & 0.124\\
        & Res9 & 4\% & 0.011 & \textbf{90\%} & \textbf{0.134} & 85\% & 0.092 \\
        & CNet & 54\% & 0.061 & \textbf{93\%} & \textbf{0.095} & 67\% & 0.087 \\
        & UNet128 & 3\% & 0.014 & \textbf{100\%} & \textbf{0.129} & 45\% & 0.052 \\
        \midrule
        \multirow{4}{*}{DD} & Res6 & 16\% & 0.031 & \textbf{100\%} & \textbf{0.139} & 96\% & 0.127 \\
        & Res9 & 5\% & 0.017 & \textbf{93\%} & \textbf{0.110} & 91\% & 0.108 \\
        & CNet & 0\% & 0.008 & \textbf{95\%} & \textbf{0.101} & 63\% & 0.087 \\
        & UNet128 & 46\% & 0.071 & \textbf{100\%} & \textbf{0.207} & 89\% & 0.089 \\
        \bottomrule
    \end{tabular}
    }
    \caption{Quantitive results of defending against attribute editing models of different settings ($\epsilon = 0.05$). Here, $\dag$ denotes the results without alternating training strategy, and $\ast$ means the gray-box setting.}
    \label{tab:black box attribute}
    \vspace{-1em}
\end{table}

\vspace{0.5em}
\noindent\textbf{Robustness in Different Adversarial Settings.} In this experiment, we consider more different adversarial settings aside from the gray-box setting.  Specifically, four types of network architectures are utilized for training target manipulation model here: a vanilla convolutional network (``CNet''), two auto-encoder like networks with 9 and 16 residual blocks respectively (``Res6'', ``Res9''), an UNet-128 network (``UNet128''). For face attribute editing task, we also leverage two type   choices of attribute domains to train $\mathbf{SM}$, namely that, the same domains ``SD'' (i.e. glasses) as training target model $\mathbf{M}$  or  the different domains ``DD'' (i.e. blond hair). Besides the gray-box setting mentioned above, we regard all the other settings as black-box setting.

\begin{table}[t]
	\footnotesize
    \centering
    \setlength{\tabcolsep}{1mm}{
    \begin{tabular}{cp{9mm}<{\centering}p{9mm}<{\centering}p{9mm}<{\centering}p{9mm}<{\centering}p{9mm}<{\centering}p{9mm}<{\centering}}
        \toprule
        \multirow{2}{*}{Setting}
        & \multicolumn{3}{p{27mm}<{\centering}}{defense} & \multicolumn{3}{p{27mm}<{\centering}}{defense$\dag$}\\
        \cmidrule(lr){2-4}\cmidrule(lr){5-7}
        & DSR & $L_1$ & $L_2$ & DSR & $L_1$ & $L_2$\\
        \midrule
        UNet256* & \textbf{100\%} & \textbf{0.057} & \textbf{0.006} & 34\% & 0.049 & 0.004\\
        CNet & \textbf{100\%} & \textbf{0.165} & \textbf{0.064} & 100\% & 0.135 & 0.060\\
        Res6 & \textbf{100\%} & 0.113 & \textbf{0.025} & 100\% & \textbf{0.115} & 0.024\\
        Res9 & \textbf{100\%} & 0.093 & 0.016 & 100\% & \textbf{0.102} & \textbf{0.019}\\
        UNet128 & \textbf{94\%} & \textbf{0.056} & \textbf{0.006} & 61\% & 0.052 & 0.005\\
        \bottomrule
    \end{tabular}
    }
    \caption{Quantitive results of defending against face reenactment models of various network structures ($\epsilon = 0.02$). Here, $\dag$ denotes the results without alternating training strategy. And $\ast$ means the gray-box setting.}
    \label{tab:black box reenactment}
    \vspace{-0.5em}
\end{table}

We can see in \Tref{tab:black box attribute} and \Tref{tab:black box reenactment} that our initiative defense of both tasks attains strong robustness in different adversarial settings. For face attribute editing task,  the defense success rate (DSR) of  the baseline method \cite{ruiz2020disrupting} degrade severely in most black-box settings, and even in gray-box setting,  DSR is only 35\%. On the other hand, our method's DSR is more than 93\% in all cases, even tested on the unseen domain in the training phase, and some visual examples are also showcased in \Fref{fig:stargan_show_1}.   

For face reenactment task, we can still realize the robustness in all adversarial settings considered above, as shown in \Tref{tab:black box reenactment}). It is worth noting that UNet performs especially well for  face reenactment task by multi-scale skip connections. In comparison, it is poorer for the performance of other network structures we adopt such as ``CNet'', ``Res6'' and ``Res9''. In our experiment, we also regard those intrinsically worse  manipulated  results as successful defense. Therefore, the defense success rate (DSR) is all 100\% in those settings even without the alternating training strategy. 
  

\subsection{Ablation study}
\begin{figure}
    \centering
    \includegraphics[width=1.0\linewidth]{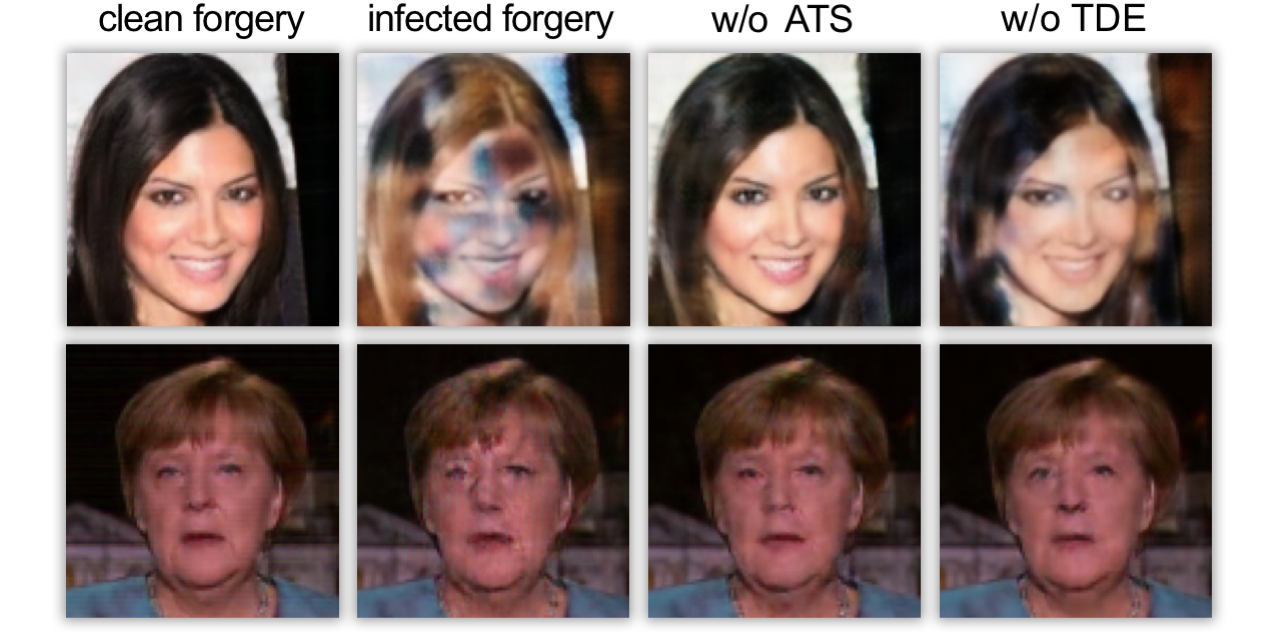}
    \caption{Visual comparison results of two tasks trained with or without alternating training and defense enhancement.}
    \label{fig:ablation}
    \vspace{-0.5em}
\end{figure}

\vspace{0.1em}
\noindent\textbf{The Importance of Alternating Training Strategy (ATS).} 
The goal of the alternating training strategy (ATS) is to evade falling into the undesired local optimum. According to the quantitative results shown in \Tref{tab:black box attribute} and \Tref{tab:black box reenactment}, the alternating training  strategy can substantially improve the defense effectiveness. Visual results is displayed in \Fref{fig:ablation}.

\vspace{0.5em}
\noindent\textbf{The Importance of Task-specific Defense Enhancement (TDE).} 
As we can find in \Fref{fig:ablation}, the more influence information absorbed in training procedure and the attention-based facial mask guidance are advantageous for corresponding tasks to enhance the defense ability. Without these, the distortion appeared on forged images might be much slighter in both defending scenarios.
%

\begin{figure}
    \centering
    \includegraphics[width=1.0\linewidth]{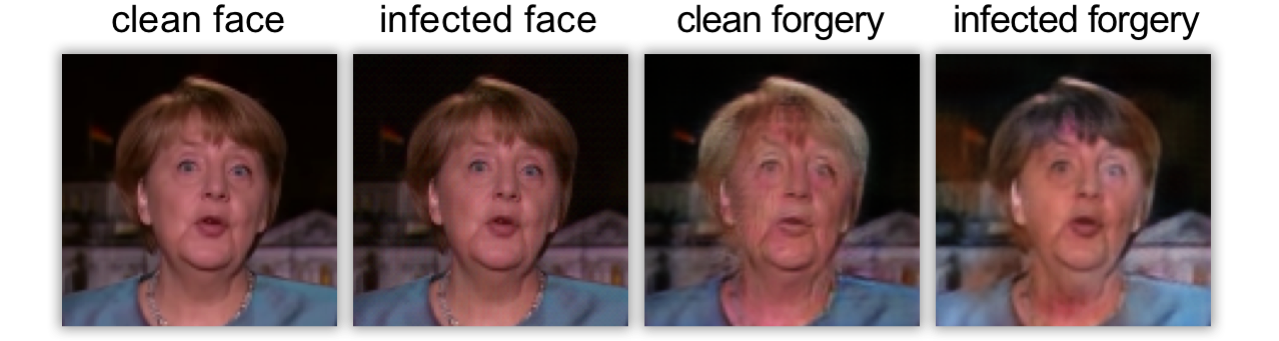}
    \caption{Visual examples of the joint defense on Merkel's speech video (wild data for face attribute editing task) and choosing gray hair as target domain (black-box setting for face reenactment task).}
    \label{fig:combine_show}
    \vspace{-0.5em}
\end{figure}

\vspace{0.5em}
\noindent\textbf{Extensions to the Joint Scenarios.} Considering that it is possible that the forger may extract one frame from the poisoned video (the second scenario) for attribute editing (the first scenario), we combine two training procedures by utilizing the first scenario's  $\mathbf{PG}$ to process each frame of videos in advance. And then we continue to process these already infected faces with the second scenario's  $\mathbf{PG}$. Based on perturbation overlaying, we can defend against the special scenes successfully while sacrificing some visual quality in this combined way, whose results can be seen in \Fref{fig:combine_show}.  

\section{Conclusion}
In this paper, we newly introduce the initiative defense against face manipulation, aiming to protect face data from being forged. By poisoning the face data, the performance of manipulation model will be degraded. To achieve this goal,  a general two-stage training framework is proposed to train a poison generator $\mathbf{PG}$. Besides, we further introduce the alternating training strategy and defense enhancement techniques into the training process. Two classic manipulation tasks, namely that, face attribute editing and face reenactment are considered. Extensive experiments demonstrate the effectiveness and robustness of our approach in different adversarial settings. Moreover, we conduct some ablation studies to prove the advantages of our design and the feasibility to the special scenarios. And we hope this work can spur more great works in this  seriously under-researched filed.


\subsection{Acknowledgments}
This work was supported in part by the Natural Science Foundation of China
under Grant U20B2047 ,U1636201, 62002334 and 62072421, Exploration Fund Project of University of Science and Technology of China under Grant YD3480002001, and by Fundamental Research Funds for the Central Universities under Grant WK2100000011. Jie Zhang is partially supported by the Fundamental Research Funds for the Central Universities WK5290000001.


\section{Ethics Statement}
With the popularity of online social platforms, people are keen to share their selfie pictures or videos. It means that plenty of human portraits are exposed in public without any protection, which provides an opportunity for malicious users to manipulate these face data. Facial manipulation may cause damage to a person's reputation, or even be used to defraud. More seriously, a manipulated political figure's video may bring threats to national security. Comparing to detection-based methods, our initiative defense framework can indeed protect against malicious usage of face data in advance rather than remedy the damages ex-post. This is effective to eliminate such risks to some extent. In general, our initiative defense will greatly inspire the relative researches in countermeasures against facial manipulation.


{
\bibstyle{aaai21}
\bibliography{references}

\begin{thebibliography}{31}
\providecommand{\natexlab}[1]{#1}
\providecommand{\url}[1]{\texttt{#1}}
\providecommand{\urlprefix}{URL }
\expandafter\ifx\csname urlstyle\endcsname\relax
  \providecommand{\doi}[1]{doi:\discretionary{}{}{}#1}\else
  \providecommand{\doi}{doi:\discretionary{}{}{}\begingroup
  \urlstyle{rm}\Url}\fi

\bibitem[{Afchar et~al.(2018)Afchar, Nozick, Yamagishi, and
  Echizen}]{afchar2018mesonet}
Afchar, D.; Nozick, V.; Yamagishi, J.; and Echizen, I. 2018.
\newblock Mesonet: a compact facial video forgery detection network.
\newblock In \emph{2018 IEEE International Workshop on Information Forensics
  and Security (WIFS)}, 1--7. IEEE.

\bibitem[{Arjovsky, Chintala, and Bottou(2017)}]{arjovsky2017wasserstein}
Arjovsky, M.; Chintala, S.; and Bottou, L. 2017.
\newblock Wasserstein Generative Adversarial Networks.
\newblock In \emph{34th International Conference on Machine Learning (ICML)},
  214--223.

\bibitem[{Bao et~al.(2018)Bao, Chen, Wen, Li, and Hua}]{bao2018towards}
Bao, J.; Chen, D.; Wen, F.; Li, H.; and Hua, G. 2018.
\newblock Towards open-set identity preserving face synthesis.
\newblock In \emph{IEEE conference on computer vision and pattern recognition
  (CVPR)}, 6713--6722.

\bibitem[{Choi et~al.(2018)Choi, Choi, Kim, Ha, Kim, and
  Choo}]{choi2018stargan}
Choi, Y.; Choi, M.; Kim, M.; Ha, J.-W.; Kim, S.; and Choo, J. 2018.
\newblock StarGAN: Unified Generative Adversarial Networks for Multi-Domain
  Image-to-Image Translation.
\newblock In \emph{IEEE Conference on Computer Vision and Pattern Recognition
  (CVPR)}.

\bibitem[{Dang et~al.(2020)Dang, Liu, Stehouwer, Liu, and
  Jain}]{dang2020detection}
Dang, H.; Liu, F.; Stehouwer, J.; Liu, X.; and Jain, A.~K. 2020.
\newblock On the detection of digital face manipulation.
\newblock In \emph{IEEE/CVF Conference on Computer Vision and Pattern
  Recognition (CVPR)}, 5781--5790.

\bibitem[{DeepFakes(2017)}]{deepfakes}
DeepFakes. 2017.
\newblock DeepFakes.
\newblock \url{http://github.com/ondyari/FaceForensics/tree/master/dataset/}.
\newblock Accessed Sept. 30, 2019.

\bibitem[{FaceSwap(2017)}]{faceswap}
FaceSwap. 2017.
\newblock FaceSwap.
\newblock \url{http://github.com/deepfakes/faceswap}.
\newblock Accessed July 4, 2019.

\bibitem[{Feng, Cai, and Zhou(2019)}]{feng2019learning}
Feng, J.; Cai, Q.-Z.; and Zhou, Z.-H. 2019.
\newblock Learning to Confuse: Generating Training Time Adversarial Data with
  Auto-Encoder.
\newblock In \emph{Advances in Neural Information Processing Systems
  (NeurIPS)}, 11971--11981.

\bibitem[{Gulrajani et~al.(2017)Gulrajani, Ahmed, Arjovsky, Dumoulin, and
  Courville}]{gulrajani2017improved}
Gulrajani, I.; Ahmed, F.; Arjovsky, M.; Dumoulin, V.; and Courville, A. 2017.
\newblock Improved Training of Wasserstein GANs.
\newblock In \emph{Advances in Neural Information Processing Systems
  (NeurIPS)}.

\bibitem[{Guo, Zhang, and Zhang(2010)}]{guo2010a}
Guo, Z.; Zhang, L.; and Zhang, D. 2010.
\newblock A Completed Modeling of Local Binary Pattern Operator for Texture
  Classification.
\newblock \emph{IEEE Transactions on Image Processing (TIP)} 19(6): 1657--1663.

\bibitem[{Han et~al.(2017)Han, Morariu, Larry~Davis et~al.}]{han2017two}
Han, X.; Morariu, V.; Larry~Davis, P.~I.; et~al. 2017.
\newblock Two-stream neural networks for tampered face detection.
\newblock In \emph{IEEE Conference on Computer Vision and Pattern Recognition
  Workshops}, 19--27.

\bibitem[{Johnson, Alahi, and Fei-Fei(2016)}]{Johnson2016Perceptual}
Johnson, J.; Alahi, A.; and Fei-Fei, L. 2016.
\newblock Perceptual losses for real-time style transfer and super-resolution.
\newblock In \emph{European Conference on Computer Vision (ECCV)}.

\bibitem[{Kim et~al.(2018)Kim, Garrido, Tewari, Xu, Thies, Niessner, P{\'e}rez,
  Richardt, Zollh{\"o}fer, and Theobalt}]{kim2018deep}
Kim, H.; Garrido, P.; Tewari, A.; Xu, W.; Thies, J.; Niessner, M.; P{\'e}rez,
  P.; Richardt, C.; Zollh{\"o}fer, M.; and Theobalt, C. 2018.
\newblock Deep video portraits.
\newblock \emph{ACM Transactions on Graphics (TOG)} 37(4): 1--14.

\bibitem[{Korshunova et~al.(2017)Korshunova, Shi, Dambre, and
  Theis}]{korshunova2017fast}
Korshunova, I.; Shi, W.; Dambre, J.; and Theis, L. 2017.
\newblock Fast face-swap using convolutional neural networks.
\newblock In \emph{IEEE International Conference on Computer Vision (ICCV)},
  3677--3685.

\bibitem[{Li et~al.(2020)Li, Bao, Zhang, Yang, Chen, Wen, and Guo}]{li2020face}
Li, L.; Bao, J.; Zhang, T.; Yang, H.; Chen, D.; Wen, F.; and Guo, B. 2020.
\newblock Face x-ray for more general face forgery detection.
\newblock In \emph{IEEE/CVF Conference on Computer Vision and Pattern
  Recognition (CVPR)}, 5001--5010.

\bibitem[{Liu et~al.(2015)Liu, Luo, Wang, and Tang}]{liu2015faceattributes}
Liu, Z.; Luo, P.; Wang, X.; and Tang, X. 2015.
\newblock Deep Learning Face Attributes in the Wild.
\newblock In \emph{IEEE International Conference on Computer Vision (ICCV)}.

\bibitem[{Matern, Riess, and Stamminger(2019)}]{matern2019exploiting}
Matern, F.; Riess, C.; and Stamminger, M. 2019.
\newblock Exploiting visual artifacts to expose deepfakes and face
  manipulations.
\newblock In \emph{2019 IEEE Winter Applications of Computer Vision Workshops
  (WACVW)}, 83--92. IEEE.

\bibitem[{Natsume, Yatagawa, and Morishima(2018)}]{natsume2018rsgan}
Natsume, R.; Yatagawa, T.; and Morishima, S. 2018.
\newblock Rsgan: face swapping and editing using face and hair representation
  in latent spaces.
\newblock \emph{arXiv preprint arXiv:1804.03447} .

\bibitem[{Nguyen, Yamagishi, and Echizen(2019)}]{nguyen2019use}
Nguyen, H.~H.; Yamagishi, J.; and Echizen, I. 2019.
\newblock Use of a capsule network to detect fake images and videos.
\newblock \emph{arXiv preprint arXiv:1910.12467} .

\bibitem[{Nirkin, Keller, and Hassner(2019)}]{nirkin2019fsgan}
Nirkin, Y.; Keller, Y.; and Hassner, T. 2019.
\newblock FSGAN: Subject agnostic face swapping and reenactment.
\newblock In \emph{IEEE international conference on computer vision (ICCV)},
  7184--7193.

\bibitem[{Pumarola et~al.(2019)Pumarola, Agudo, Martinez, Sanfeliu, and
  Moreno-Noguer}]{Pumarola_ijcv2019}
Pumarola, A.; Agudo, A.; Martinez, A.; Sanfeliu, A.; and Moreno-Noguer, F.
  2019.
\newblock GANimation: One-Shot Anatomically Consistent Facial Animation.
\newblock \emph{International Journal of Computer Vision (IJCV)} .

\bibitem[{Qian et~al.(2020)Qian, Yin, Sheng, Chen, and Shao}]{qian2020thinking}
Qian, Y.; Yin, G.; Sheng, L.; Chen, Z.; and Shao, J. 2020.
\newblock Thinking in Frequency: Face Forgery Detection by Mining
  Frequency-aware Clues.
\newblock \emph{arXiv preprint arXiv:2007.09355} .

\bibitem[{Ronneberger, Fischer, and Brox(2015)}]{ronneberger2015u-net:}
Ronneberger, O.; Fischer, P.; and Brox, T. 2015.
\newblock U-Net: Convolutional Networks for Biomedical Image Segmentation.
\newblock \emph{MICCAI} 234--241.

\bibitem[{R{\"o}ssler et~al.(2019)R{\"o}ssler, Cozzolino, Verdoliva, Riess,
  Thies, and Nie{\ss}ner}]{rossler2019faceforensics++}
R{\"o}ssler, A.; Cozzolino, D.; Verdoliva, L.; Riess, C.; Thies, J.; and
  Nie{\ss}ner, M. 2019.
\newblock Faceforensics++: Learning to detect manipulated facial images.
\newblock \emph{arXiv preprint arXiv:1901.08971} .

\bibitem[{Ruiz, Bargal, and Sclaroff(2020)}]{ruiz2020disrupting}
Ruiz, N.; Bargal, S.~A.; and Sclaroff, S. 2020.
\newblock Disrupting Deepfakes: Adversarial Attacks Against Conditional Image
  Translation Networks and Facial Manipulation Systems.
\newblock \emph{arXiv:2003.01279} .

\bibitem[{Shan et~al.(2020)Shan, Wenger, Zhang, Li, Zheng, and
  Zhao}]{shan2020fawkes}
Shan, S.; Wenger, E.; Zhang, J.; Li, H.; Zheng, H.; and Zhao, B.~Y. 2020.
\newblock Fawkes: Protecting Personal Privacy against Unauthorized Deep
  Learning Models.
\newblock In \emph{Proceedings of USENIX Security}.

\bibitem[{Thies et~al.(2016)Thies, Zollhofer, Stamminger, Theobalt, and
  Niebner}]{thies2016face2face:}
Thies, J.; Zollhofer, M.; Stamminger, M.; Theobalt, C.; and Niebner, M. 2016.
\newblock Face2Face: Real-Time Face Capture and Reenactment of RGB Videos.
\newblock In \emph{IEEE Conference on Computer Vision and Pattern Recognition
  (CVPR)}.

\bibitem[{Wu et~al.(2018)Wu, Zhang, Li, Qian, and
  Change~Loy}]{wu2018reenactgan}
Wu, W.; Zhang, Y.; Li, C.; Qian, C.; and Change~Loy, C. 2018.
\newblock Reenactgan: Learning to reenact faces via boundary transfer.
\newblock In \emph{European conference on computer vision (ECCV)}, 603--619.

\bibitem[{Yu et~al.(2018)Yu, Wang, Peng, Gao, Yu, and Sang}]{yu2018bisenet:}
Yu, C.; Wang, J.; Peng, C.; Gao, C.; Yu, G.; and Sang, N. 2018.
\newblock BiSeNet: Bilateral Segmentation Network for Real-Time Semantic
  Segmentation.
\newblock In \emph{European Conference on Computer Vision (ECCV)}, 334--349.

\bibitem[{Zhang et~al.(2010)Zhang, Gao, Zhao, and Liu}]{zhang2010local}
Zhang, B.; Gao, Y.; Zhao, S.; and Liu, J. 2010.
\newblock Local Derivative Pattern Versus Local Binary Pattern: Face
  Recognition With High-Order Local Pattern Descriptor.
\newblock \emph{IEEE Transactions on Image Processing (TIP)} 19(2): 533--544.

\bibitem[{Zhang et~al.(2018)Zhang, Isola, Efros, Shechtman, and
  Wang}]{zhang2018perceptual}
Zhang, R.; Isola, P.; Efros, A.~A.; Shechtman, E.; and Wang, O. 2018.
\newblock The Unreasonable Effectiveness of Deep Features as a Perceptual
  Metric.
\newblock In \emph{IEEE conference on computer vision and pattern recognition
  (CVPR)}.

\end{thebibliography}
}
\end{document}